\documentclass[11pt,a4paper]{article}
\usepackage[hyperref]{emnlp2020}
\usepackage{times}
\usepackage{latexsym}

\usepackage{booktabs}
\usepackage{graphicx}
\usepackage{todonotes}
\usepackage{amsmath}
\usepackage{linguex}
\usepackage{soul}
\usepackage{microtype}
\usepackage{tikz}
\usepackage{tikz-dependency}
\usetikzlibrary{decorations.pathreplacing}

\def\checkmark{\tikz\fill[scale=0.4](0,.35) -- (.25,0) -- (1,.7) -- (.25,.15) -- cycle;}

\aclfinalcopy 

\title{WikiGUM: Exhaustive Entity Linking for Wikification in 12 Genres}

\author{Jessica Lin \\
   Department of Linguistics \\
  Georgetown University \\
  \texttt{yl1290@georgetown.edu} \\\And
  Amir Zeldes \\
   Department of Linguistics \\
  Georgetown University \\
  \texttt{amir.zeldes@georgetown.edu} \\}

\date{}

\begin{document}
\maketitle


\begin{abstract}
Previous work on Entity Linking has focused on resources targeting non-nested proper named entity mentions, often in data from Wikipedia, i.e. Wikification. In this paper, we present and evaluate WikiGUM, a fully wikified dataset, covering all mentions of named entities, including their non-named and pronominal mentions, as well as mentions nested within other mentions. The dataset covers a broad range of 12 written and spoken genres, most of which have not been included in Entity Linking efforts to date, leading to poor performance by a pretrained SOTA system in our evaluation. The availability of a variety of other annotations for the same data also enables further research on entities in context. 
\end{abstract}

\section{Introduction}

Entity linking (EL) involves identifying entities within a text and subsequently linking their mentions to a knowledge base or table of authorities. The former step is often referred to as Named Entity Recognition (NER) and the latter may also be referred to as entity disambiguation. In this study, we will focus on the latter task by following the popular approach of mapping named entities to Wikipedia entities \cite{milne2008aquaint,shnayderman2019fast}, i.e.~Wikification. 

Wikification is the task of adding links to Wikipedia pages to mentions of named entities in a written or spoken text. This task supports Natural Language Understanding in downstream tasks such as question answering, summarization, and relation extraction. However the scope and structure of EL depends heavily on datasets which are either automatically derived from hyperlinked text and thus suffer some limitations, or are created via human annotation, a time-consuming and expensive task. Despite numerous existing EL datasets \cite{cucerzan2007msnbc,ji2015tackbp15,kulkarni2009iitb,milne2008aquaint,ratinov2011ace}, few  have attempted to capture \textit{nested} entity structure, as in Figure \ref{fig:nested}, which never occurs in hyperlinks, which cannot be nested. 



\begin{figure}[h!tb]
\centering
\includegraphics[width=7cm, trim={0cm 15cm 24cm 0cm}, clip]{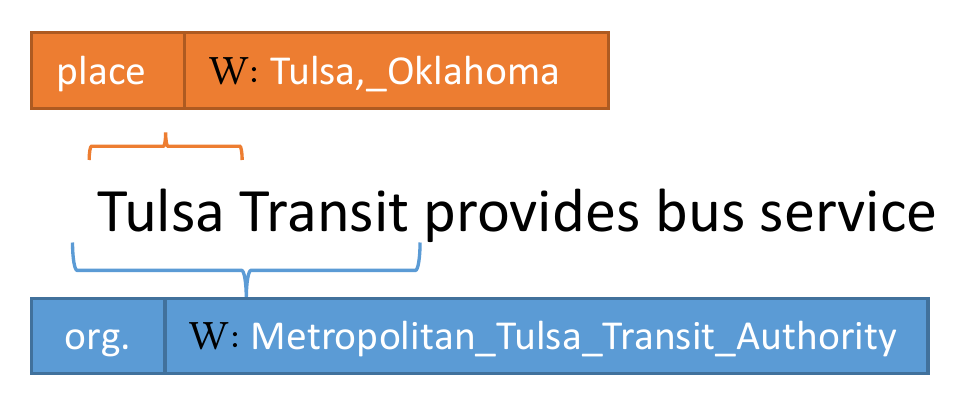}
\caption{Nested entity linking.}
\label{fig:model_architecture}
\par\vspace{-10pt}\par
\end{figure}\label{fig:nested}

Instead, annotations have focused on flat mention structure from popular online sources, leaving out important information in nested entities that can be useful for downstream tasks. Closest to the resource presented here is the Nested Named Entities (NNE) dataset \cite{ringland2019nne}, which is a large, manually-annotated, nested named entity dataset over English newswire, however, it does not include entity linking. Although NNE includes fine-grained semantic information in nested entity types, it is not linked to any identifiers (e.g.~a Wikipedia page).
Furthermore, even if used for mention recognition, the data is not ideal for testing on diverse genres, as NNE solely covers news text. We also note other datasets capturing some nested entity structure, such as the Abstract Meaning Representation (AMR) corpus \citep{banarescu2013amr}, which includes compositional nesting e.g.~in possessives such as \textit{Toronto's international airport}, composed of a city and an airport. However, since AMR is not word-aligned to text, even those nested entities that are covered are not aligned to their textual position. \par

In this paper, we present and evaluate a gold standard wikified dataset, called WikiGUM, in which named and non-named entities have been annotated manually. WikiGUM is based on the existing GUM dataset (Georgetown University Multilayer corpus, \citealt{zeldes2017gum}), and goes beyond other EL corpora, in covering \textit{all} mentions of named entities (NEs), including non-named and pronominal mentions, as well as nested mentions, for 12 genres of English text. WikiGUM also enables assessment of EL annotations by highlighting challenges that are common in our dataset, and reveals the relatively poor coverage of state-of-the-art NLP systems for EL in diverse genres (Section \ref{sec:eval}). Taken together, we aim to facilitate new research on nested NER and EL, to promote recognition of all NE mentions and a deeper understanding of the hierarchical structure of entities in text.

\section{WikiGUM}
\label{sec:2}
The underying GUM \cite{zeldes2017gum} corpus is a manually annotated dataset with multiple layers, including POS tagging (Penn tags, CLAWS5, Universal POS), sentence types (e.g.~declarative, imperative, yes/no question), UD dependency trees \cite{nivre2016ud}, coreference resolution (including bridging anaphora and split antecedents), and RST discourse parses \cite{mann1988rst}. Data covers 12 genres: academic, biographies, conversation, fiction, forums, how-to, interviews, news, speeches, textbooks, travel and vlogs. \par

\begin{table*}[htbp]
\centering
\resizebox{0.9\textwidth}{!}{%
\begin{tabular}{@{}lllllll@{}}
\toprule
\textbf{Text Types} & \textbf{Source} & \textbf{Documents} & \textbf{\# of NE Mentions}& \textbf{\# of Nested Wikified Mentions}& \textbf{Total Mentions} & \textbf{Tokens} \\ \midrule
Interviews & Wikinews & 19 & 1,146 & 107 & 5,204 & 18,037 \\
News stories & Wikinews & 21 & 1,221 & 217 & 4,130 & 14,094 \\
Travel guides & Wikivoyage & 17 & 1,327 & 174 & 4,087 & 14,955 \\
How-to guides & WikiHow & 19 & 94 & 8 & 4,469 & 16,920 \\
Academic writing & various & 16 & 329 & 48 & 4,486 & 15,110 \\
Biographies & Wikipedia & 20 & 2,450 & 413 & 5,763 & 17,951 \\
Fiction & various & 18 & 195 & 9 & 4,737 & 16,307 \\
Forum discussions & Reddit & 18 & 196 & 2 & 4,530 & 16,286 \\
Conversations & UCSB corpus & 5 & 31 & 0 & 1,477 & 5,698 \\
Political speeches & various & 5 & 316 & 32 & 1,423 & 4,831 \\
CC Vlogs & YouTube & 5 & 39 & 1 & 1,355 & 5,180 \\
Textbooks & OpenStax & 5 & 188 & 21 & 1,507 & 5,376 \\
\hline
\multicolumn{2}{l}{Total} & 168 & 7,352 & 1,032 & 43,168 & 150,745 \\ \bottomrule
\end{tabular}%
}
\caption{Statistics on WikiGUM}
\label{tb:1}
\end{table*}

WikiGUM adds a layer of Wikipedia identifiers to all NEs in GUM, which are identified automatically by having the gold PTB POS tag NNP(S) for their syntactic head, based on gold syntax trees (for some resulting issues, see below), as well as non-named mentions coreferring to them based on coreference annotations. Since GUM is expanded by students in classroom annotation every year, and we plan to continue adding Wikification in the future, no closed or pre-prepared ontology is applied to the Wiki identifiers, making the task simpler for student annotators who only need to find a corresponding Wikipedia article.

That said, the existing 10 entity types in GUM\footnote{Types: \textsc{person, place, org, animal, plant, event, time, substance, abstract} and inanimate \textsc{object}.} mean that our EL benefits from the same categorization scheme as a rough ontology, and the availability of semantic information from WikiData means that many relationships between entities can be explored. All referential NPs, including pronouns and even clauses (if they co-refer with a named entity based on GUM's coreference annotations, for example movie titles), were selected as markables for annotation. Note that nested markables are always included, for example: 

\ex. [\textit{the airport in} [\textit{Cuba}]\textsubscript{place}]\textsubscript{place}

Our general guideline for entity linking is that NEs, including  pronominal and non-named mentions, were manually linked to the corresponding Wikipedia article whenever one exists, using the version controlled online editor GitDox \cite{zhang2017gitdox}. For example: 

\ex. \textit{Kim likes} [\textit{The Terminator}]\textsubscript{abstract}.~[\textit{This movie}]\textsubscript{abstract} \textit{is her favorite}.

In this example, the span \textit{This movie} should also be linked to the Wikipedia page that refers to \textit{The Terminator (the movie)}. Statistics on WikiGUM, which is freely available under the same Creative Commons license as GUM, are shown in Table~\ref{tb:1}.\par

Although the basic Wikification task is fairly straightforward, some ambiguous/tricky cases during annotation included:

\begin{itemize}
  \item \textbf{Generic terms}: some capitalized common nouns that have Wikipedia links appear within NEs, and are tagged NNP(S), but do not correspond to named entities. For example, \textit{Oil} is incorrectly proposed as a NE due to capitalization within the NE \textit{the Oil Capital of the World} (referring to Tulsa, OK) and due to the POS tag NNP. It can be tempting to link `oil' as a NE candidate to the Wikipedia article `Petroleum'. However in context, `oil' is a generic, non-named modifier to `Capital', and should not be linked as a NE. Annotators should be mindful of context of terms tagged NNP(S) within NEs, rather than linking any NNP span.
  \item \textbf{Subset of entity with the same type}: a common type of ambiguity  for \textsc{place} entities arises when names are reused in different countries, regions, cities or villages. For example, terms like `North', `South', `East', and `West' as a subset of a region are hard to disambiguate, and they are common in street names in North America. In this case, annotators must look at the broader context and carefully check whether the entity refers to a subset or not, for example cities and their metropolitan areas, streets with and without cardinal directions, or other parts of cities which sometimes have separate Wikipedia entries.
  \item \textbf{Distinct links for identical mentions}: It is sometimes hard for annotators to realize that an entity string has several EL variants. This happens often in abbreviations, which may be labelled with the wrong entity type. For example, `JFK' can be a person's name (the 35\textsuperscript{th} US President) or a place name (JFK Airport in New York), depending on context. We instructed annotators to prioritize the existing entity type annotation: if JFK is tagged as a place, it is linked to the article about the airport. Another common issue affects ancient place names which do not exist nowadays, resulting in difficulty for EL. For instance, \textit{Jorvik} is the viking name of York, and was therefore linked to the closest equivalent article, `Scandinavian York' rather than `York'. In other cases, we relied on the coreference annotations to establish equivalence: for example \textit{England's City of Festivals} was labeled as coreferring with \textit{York}, and was therefore considered equivalent to York for EL purposes.
  
  \item \textbf{Lack of background knowledge}: In some cases context alone cannot help annotators decide on the right sense of an entity, especially in academic texts, but also in discussion forums. Academic articles often assume readers have detailed knowledge of the topic and thus provide little context for the target entity. For example, `Su' in `Su et al. 2016' is a named entity, but it may be difficult to know whether there is a corresponding Wikipedia article based solely on the author's name.
\end{itemize}

\section{Related work}

Table~\ref{tb:2} compares WikiGUM to other EL corpora. Most current EL datasets are based on newswire text, overlooking the impact of genre on EL -- for example, \citet{dai2018biomed} notes that the biomedical domain involves complex and unique entity mentions. As EL datasets are developed for evaluation of EL systems, out-of-domain data could create challenges for conventional tools. Furthermore, most previous work \cite{cucerzan2007msnbc,ji2015tackbp15, kulkarni2009iitb,milne2008aquaint,ratinov2011ace} has focused on identifying and classifying atomic, flat mention structures, leaving out the semantic information available in nested mentions. 

As shown in Table~\ref{tb:2}, most datasets do not contain nested entity linking annotations, with ACE2004 being the exception \cite{ratinov2011ace}. Unfortunately, no dataset covers all mentions of named entities, including their non-named mentions (`the same airport', or `it'). 
As mentioned above, we do see some research on nested entity structure \cite{glavavs2014constructing,hong2016building}, for example the NNE corpus  \cite{ringland2019nne} contains fine-grained semantic information including e.g.~the category \textsc{city} nesting a \textsc{state}, which could easily be used for EL. However they are not disambiguated or linked to a table of authorities, in addition to excluding non-named mentions of the same entities. WikiGUM thus differs from previous EL datasets and is rich in terms of both genre and entity structure, as well as being among the larger available datasets as shown in the Table.

\begin{table*}[htbp]
\centering
\resizebox{\textwidth}{!}{%
\begin{tabular}{@{}llrrrlll@{}}
\toprule
Dataset & Paper & \# of documents & \# of NEs & \# of genres & NE/N & Pronouns & Nested Entities \\ \midrule
WikiGUM & -- & 168 & 7,352 & 12 & NE\&N & \checkmark & \checkmark \\
ACE2004 & \cite{ratinov2011ace} & 36 & 256 & 1 & NE &  & \checkmark \\
AIDA-A & \cite{hoffart2011aida} & 216 & 5,917 & 1 & NE &  &  \\
AIDA-B & \cite{hoffart2011aida} & 231 & 5,616 & 1 & NE &  &  \\
AQUAINT & \cite{milne2008aquaint} & 50 & 727 & 1 & NE\&N &  &  \\
Derczynski & \cite{derczynski2015tweet} & 182 & 210 & 1 & NE &  &  \\
IITB & \cite{kulkarni2009iitb} & 107 & 17,200 & 1 & NE &  &  \\
KORE50 & \cite{hoffart2012kore50} & 50 & 148 & 1 & NE &  &  \\
MSNBC & \cite{cucerzan2007msnbc} & 20 & 656 & 1 & NE &  &  \\
n3-RSS-500 & \cite{roder2014n3} & 500 & 1,000 & 1 & NE &  &  \\
n3-Reuters-128 & \cite{roder2014n3} & 128 & 880 & 1 & NE &  &  \\
OKE2015 & \cite{nuzzolese2015oke} & not specified & 718 & 1 & NE\&Roles &  &  \\
OKE2016 & \cite{nuzzolese2016oke} & not specified & 940 & 1 & NE\&Roles &  &  \\ \bottomrule
\end{tabular}%
}
\caption{English EL datasets. NE/N indicates whether only named entities (NE) or also common  nouns (N) are included. Note that ACE 2004 is a subset of the documents used in the ACE 2004 coreference documents.}
\label{tb:2}
\end{table*}

\section{Evaluation}\label{sec:eval}

In this section we evaluate inter-annotator agreement, as well as the extent to which existing Wikification technology already captures the information in WikiGUM.

\paragraph{Inter-annotator agreement} Measuring agreement for Wikification involves two main complementary aspects: span detection and Wikification (including the decision whether to link an entity and to what). Since GUM already contains mention boundaries and named/non-named status, we focus on the latter task, measuring linking agreement. To calculate agreement, we carried out an inter-annotator agreement experiment by double annotating 3,103 tokens of corpus data containing 237 entities after adjudication, about 3\% of the data. We compute both Cohen's Kappa and simple percent agreement (percentage of exact match), shown in Table~\ref{tb:3}. Note that computing Cohen's Kappa here is somewhat artificial, as in the real world there is an (almost) unlimited space of possible Wikipedia identifiers. For simplicity, we define the space of possible links as the union of any values annotators used in this subset, meaning that any link chosen at any point by any annotator is considered a possible value for the annotation, and any disagreement is penalized by the metric.\footnote{An anonymous reviewer has asked whether this means that search ambiguity and an overwhelming number of options impacted our process: this is certainly true, and somewhat inevitable given that annotators were unrestricted in the identifiers they could choose from Wikipedia. However the high level of absolute agreement suggests that in practice annotators were surprisingly internally consistent.}

\begin{table}[h!tb]
\begin{center}
\begin{tabular}{ l l } 
 \hline
 Metric & Score \\ 
 \hline
 Agreement & 0.8903  \\
 Cohen's $\kappa$ & 0.8782 \\
 \hline
\end{tabular}
\caption{Results of the inter-annotator agreement experiment}
\label{tb:3}
\end{center}
\end{table}

Results in Table~\ref{tb:3} show that agreement is far beyond chance, with Kappa=.87 and simple agreement of .89. While this indicates very good agreement, raters did disagree on ambiguous cases, which is worth discussing. A major source of disagreement involves linking the same entity string to distinct but related identifiers, i.e.~the \textbf{name variants} issue highlighted in Section~\ref{sec:2}. This is often due to lack of context information: for example, in the sentence ``\textit{CC makes things more complex}'', the mention \textit{CC} could be linked to \textit{Creative Commons license} (public copyright license) or \textit{Creative Commons} (organization that produced the Creative Commons license). In this case, broader context and reasoning are required to make consistent decisions. Another example is the sentence ``\textit{According to the Arts and Humanities Citation Index Professor Chomsky is the eighth most cited scholar of all time.}'', the mention \textit{Arts} is not a NE and should not be linked, but was linked by one annotator to ``\textit{The arts}'', which has a linkable article and can easily be confused with a named concept of sorts. 




\paragraph{NLP coverage} 

To evaluate the usefulness of WikiGUM beyond existing resources, we test a recent SOTA pretrained end-to-end neural system (\textit{e2e}, \citealt{kolitsas-etal-2018-end}) on the test set and compare it to a baseline strategy. Our baseline system uses a neural constituent parser \cite{mrini-etal-2020-rethinking} to identify predicted noun phrases and simply checks the exact string of every phrase headed by a proper noun to see if it has a Wikipedia article (using the Python library \texttt{wikipedia}). Since the SOTA system cannot identify nested mentions, but we do not know which one of two nested mentions it might identify (the bigger or smaller one), we evaluate in multiple scenarios: counting all mentions, only unnested mentions, and, since the system cannot identify pronouns, with and without them.

\begin{table}[h!bt]
\normalsize
\begin{tabular}{llrrrr}
\hline
 & data & P & R & F1 & links \\
\hline
\textit{e2e} & all & 0.398 & 0.192 & 0.259 & 827  \\
 & -pron & 0.441 & 0.259 & \textbf{0.327} & 677  \\
 & -nest & 0.363 & 0.203 & 0.260 & 713  \\
 & -both & 0.363 & 0.253 & 0.298 & 573  \\
 \hline
\textit{baseline} & all & 0.480 & 0.182 & 0.264 & 827  \\
 & -pron & 0.480 & 0.223 & \textbf{0.304} & 677 \\
 & -nest & 0.363 & 0.203 & 0.260 & 713 \\ 
 & -both & 0.391 & 0.214 & 0.277 & 573
\\ \hline 
\end{tabular}
 \caption{Baseline and e2e results on WikiGUM test set.}
\end{table}\label{tab:nlp}

The results in \ref{tab:nlp} show that \textit{e2e}, even when trained on the largest available Wikification dataset (AIDA, $\approx$1,000 documents, 18K links) does not generalize well to the domains found in our corpus, barely outperforming a naive lookup baseline. Comparing the scenarios, we see that best performance for both systems is achieved when removing pronouns, which is unsurprising since neither strategy can be expected to link them. However removing nested mentions does not result in higher scores: this is because some common targets, such as places, are often nested in larger names (organizations, office-holders), and removing them disrupts score gains from their correct identification. Nevertheless, the most lenient possible evaluations are in the low 30s, as opposed to a score of 82.6 on AIDA \cite[524]{kolitsas-etal-2018-end}. This suggests that, unsurprisingly, the corpus covers a range of entities and contexts that are under- or unrepresented in previous benchmarks.

\section{Conclusion}

This paper presented WikiGUM, the first exhaustive Wikification dataset for named entity linking, including nested and pronominal mentions in 12 genres of English text. Our evaluation suggests a high level of agreement, as well as coverage for a significant amount of entities not retrieved by a SOTA neural linking system. We hope that this dataset will enable further research on entity linking and increase coverage for all types of linkable named entities across a broad spectrum of genres, both spoken and written.

\bibliographystyle{acl_natbib}
\bibliography{emnlp2020}


\end{document}